# Vehicle trajectory prediction in top-view image sequences based on deep learning method

Zahra Salahshoori nejad[1*], Hamed Heravi[1*], Ali Rahimpour Jounghani[2], Abdollah Shahrezaie[1], Afshin Ebrahimi[1].

*same contribution

[1]Department of Electrical Engineering, Sahand University of Technology, Tabriz, Iran

[2]Department of Psychological Sciences, University of California, Merced, California, USA

**Type of article:** Research paper

**Corresponding Author:**

Afshin Ebrahimi

E-mail: aebrahimi@sut.ac.ir

Address:

Tel: 098 4133459374

Fax: 098 4133459374

## Abstract

Annually, a large number of injuries and deaths around the world are related to motor vehicle accidents. This value has recently been reduced to some extent, via driver-assistance systems. Predicting the trajectory is influenced by numerous factors, such as drivers' behavior, history of the vehicle's movement and its surrounding vehicles, and their position on the traffic scene. Moreover, to predict automated vehicles' path, a model with low computational complexity is proposed, which is trained by images taken from the road's aerial image. Our method is based on an encoder-decoder model utilizing a social tensor to model the effect of the surrounding vehicles' movement on the target vehicle. The proposed model can predict the vehicle's future path in freeway by observing the images related to the history of the target vehicle's movement and its neighbors. Deep learning was used as a tool for extracting the features of these images. Using the HighD database, an image dataset of the road's aerial image was created, and the model's performance was evaluated on this new database. We achieved the RMSE of 1.91 for the next 5 seconds and found that the proposed method had less error than the best path-prediction methods in previous studies.

## 1. Introduction

Driving is a highly skilled task that requires an extensive understanding of the other road users' goals. This understanding allows drivers to take a safe path through the traffic. Although this issue may be natural for an experienced human driver, an accurate understanding of the other drivers' goals remains an unsolved problem for advanced driver-assistance systems and automated vehicles.

While driving, drivers must continuously interact with one another. Their behavior considerably affects the safety of driving. Notably, an experienced driver takes appropriate measures based on accurate predictions of the traffic status [1]. This specific phenomenon is also performed in automated driving. Automated vehicles must ensure the driver's safety and comfort when passing through the traffic flow. Reliable prediction of vehicles' future movements must be performed quickly and considering the traffic components [2]. To this end, the vehicle should be able to analyze the future movement of the surrounding vehicles. This ability has been included in the existing path-design algorithms [3, 4], which depend on estimating the future paths of the surrounding vehicles [5]. If the vehicle can achieve a proper understanding of its surrounding environment, it can design its path of movement in proportion to the situation at hand, such that the risk of accidents would be minimized.

To improve the safety of automated driving, it is essential to use a reliable prediction method. This method must provide accurate prediction in sufficient time and should have satisfactory efficiency and reliability.

Generally, two steps are required for designing such a practical system. In the first step, the immediate identification and tracking of the surrounding vehicles and, in the second step, the design and execution of algorithms with low computational complexity for ensuring real-time deduction should be performed because timely and successful prediction can lead to better coordination between automated vehicles [6].

One of the most important parts of designing a path-prediction algorithm is to select appropriate inputs for model training. The proposed methods need the manual extraction of features to train prediction models. As large driving datasets are required, the non-use of automatic data labeling methods that can accelerate real-time applications' functioning is a literature gap. One solution is to directly utilize images to view the driving scene for the path prediction and programming algorithms that do not need multiple feature engineering steps in many methods.

Recently, deep convolutional neural networks (CNNs) have become widely popular in many domains and can learn powerful hierarchical features from complex inputs. They are among the most well-known and successful deep neural networks that can offer the best performance as feature extractors in information extraction and input processing. Convolutional layers directly receive time and place features from the input and eliminate the need for feature engineering.

From among different deep neural network structures, recurrent neural networks (RNNs) are widely used for learning temporal dynamics in time series data. Long short-term memories (LSTMs) are a specific component of RNNs that can learn the long-term relationships among features. They consider the input information as intervals and model their dependencies [7].

Since they access the entire history of time series values using the recurrent capability, LSTMs are regarded as the most advanced time series prediction methods. These methods receive information from the history of movement paths in previous frames and produce a set of target coordinates for the next frames [8]. Therefore, different methods using LSTM have been proposed for solving path-related problems. The input of a prediction model and its expected output are a sequence of temporally related data to one another. Therefore, the proposed model must create temporal relations in input and output data between them [9].

In this study, an encoder-decoder architecture and an extra module are used for modeling the spatial interactions among the surrounding vehicles. This module is a new type of the social pooling approach introduced by Alahi et al. [10] for modeling the interaction between pedestrians near one another, developed by Deo and Trivedi [11] learning the mutual dependence of vehicles. In the encoder part, convolutional layers on each part of the input image have been used to automatically extract the traffic scene's local features and record the correlations between time steps from the input images. The extracted features create a social tensor fed to a convolutional neural network to learn mutual local dependence. Also, an LSTM decoder performs the prediction for multiple future time horizons. The goal of enc-dec is to model the conditional probability of the output sequence on the condition of having the input sequence.

The proposed method attempts to simplify the network architecture and directly use the road structure by using the CNN capability in feature extraction. This system can generate path sequences from the camera's raw input without needing complex sensors such as LiDAR. As the proposed architecture does not have the complexity of the previously introduced methods, it can compete with the existing modern methods by providing shorter deduction time.

## 2. Related Research

In recent years, vehicle path prediction has become a hot research topic and has been regarded as a critical element in many studies, including those in the field of robotics and automated vehicles. Many studies have attempted to predict the behaviors and paths of pedestrians and vehicles on roads [38, 39]. In recent years, various approaches have been proposed for this purpose, each of which has its features.

Lefevre et al. [12] categorized vehicle behavior prediction models into physics-based, maneuver-based, and interaction-aware models. Physics-based models have the lowest degree of complexity and are limited to short-term and unreliable predictions. Also, they completely ignore drivers' tendencies, which reduces their reliability.

Maneuver-based models resolve this problem. In these models, each driver's movements are taken as a maneuver; this is done independently from other traffic components. Generally, a maneuver-based prediction is more reliable in the long term. However, if the feature space dimensions are increased, the categorization problem becomes markedly more difficult.

Interaction-aware models are the complete models proposed so far because they can provide longer-term predictions than physics-based models and are more reliable than maneuver-based models since they consider the dependency between vehicles [40, 41]. These models view vehicles as maneuvering objects that interact with one another. In other words, it is assumed that the movement of one vehicle is affected by the other vehicles in the same environment. This event leads to a better understanding of the current status and better evaluation of the future [42,43,44,45,46]. Although interaction awareness features are necessary for maneuver prediction, few approaches have existed about considering these features before 2014. The number of studies using these features, explicitly or implicitly, has increased in recent years.

Multiple advanced approaches, mostly based on deep learning, have been proposed since 2014 and can be classified based on the input type [13].

First, we will separate the vehicles in the scene as follows: 1) Target vehicle (TV): The vehicle, the path of which is predicted;2) Ego vehicle (EV): The vehicle, on which the movement information recording and measuring sensor is mounted;3) Surrounding vehicles (SVs): Vehicles located in the neighborhood of the TV that affect the act of prediction.

Sensors, radars, or cameras usually record the data used for the path prediction problem. In one classification, the existing studies can be divided into the following classes based on the type of input data: The track history of the TV, the track history of the TV and its neighbors, aerial view, and bird's-eye view. The latter two classes can be regarded as sub-classes of interaction-aware approaches.

**1) Track history of TV:** The conventional approach in these models for TV behavior prediction is to use its current status (e.g., position, speed, acceleration, direction) or the path followed over time. In [14, 15, 16], position tracking, speed, and direction of the TV in previous seconds have been used for predicting its behavior. All these studies examine the behavior of the TV without any neighbor in the environment. Some deep learning-based methods use this type of input for predicting vehicle behavior in a driving setting in the presence of other vehicles.

Xin et al. [17] showed that, due to limitations of the data recording sensor and the problem of occlusion, the information on some SVs is not available, even though some neighboring vehicles are always within the view of the EV's sensor. Since vehicles' interaction affects the prediction, some SVs' invisibility can negatively affect the precision of the prediction.

Although the TV's movement history contains useful information on its short-term future movement, the use of the vehicle's movement history alone cannot lead to accurate results, especially for long-term prediction in busy driving settings.

**2) Track history of TV and SVs:** One approach for considering the interaction between vehicles is using the track history of the TV and SVs as the input to the prediction model. Like the TV states, SVs can be estimated in the object detection module in the EV. At any rate, some SVs may be outside the reach of the EV or hidden from view by the other vehicles on the path.

Studies on the classification of vehicles in the scene into SVs and NVs differ from each other. In [18, 19, 20], the TV's status history and six nearby SVs have been used to predict the TV's path. The authors of [21, 22] regarded the vehicles in three lanes around the TV as its SVs, including two vehicles in front. They showed that taking into account more vehicles in the input data can improve behavior prediction performance. In [23, 24, 25], instead of assuming a constant number of vehicles as SVs, an area was defined as the neighboring range. This means that the interactions of the vehicles in this range were used in the prediction model. Most of these studies assume that all the SVs' states are always visible, not the realistic assumption in automated driving applications. A more realistic approach should always consider the path data-recording sensor's disorders, such as occlusion and noise.

**3) The simplified bird's-eye view:** An alternative for considering the vehicles' interactions is using the simplified bird's-eye view (BEV) of the environment. In this method, stable and dynamic objects and the other elements in the environment are usually shown with a set of polygons and lines in a BEV image. The result is an image similar to a map that maintains the objects' size and location (e.g., vehicles) and the road's geometry while ignoring their texture.

Lee et al. [26] combined the camera and radar data to create a bi-channel BEV image. One channel specifies whether each pixel contains vehicles, and the other channel determines the existence of lane signs. For the past n frames, these images are created and combined to generate a 2n-channel image as the input to the prediction model. Instead of using a sequence of binary images, a single BEV image has been used in [27, 28] to demonstrate objects' existence over time. In this image, each element of the scene (e.g., road, sidewalk) loses its real texture and, instead, has a specific color based on its meaning. Vehicles are shown with colored boxes, and their movement history is indicated with the same color but darker.

To enrich the temporal information in the BEV image, Deo and Trivedi [11] used a social tensor, which was first introduced in [10] (known as the social pooling layer). A social tensor is a grid around the TV, and the cells in each grid of this network are filled with the processed temporal data (LSTM hidden state) of the vehicle. Therefore, social tensor includes the vehicle's temporal dynamics and mutual spatial dependency.

The cited references did not consider the defects of the sensor in depicting the inputs. A dynamic occupancy grid map (DOGMa) has been employed to resolve this issue [29, 30, 31]. DOGMa is created by combining the data of different sensors and presents a BEV image of the environment. The channels in this image include the probability of occupation and an estimate of speed for each pixel. The speed information helps differentiate the stable and dynamic objects in the environment.

Nevertheless, it does not provide complete information on the history of dynamic objects.

The first advantage of a simple BEV is that it is flexible in terms of the type of representation's complexity. Therefore, it can adapt itself to applications with different computational limitations. Also, a combination of different data from various sensors may be converted into a single BEV image.

In addition to the inherent challenges of vehicle behavior prediction, executing a behavior prediction module in automated vehicles has several practical restrictions. For instance, there are limited computational resources for on-board execution in automated vehicles. Moreover, these vehicles cannot completely view the environment due to limitations in their on-board sensors.

The majority of studies have assumed they can access an extensive top-to-down view without obstacles to the driving environment accessible by infrastructural sensors. These data can be accessed if there is a communication channel with sufficient capacity between the infrastructures and the automated vehicle [13].

Using the BEV images obtained from aerial cameras is the right choice for preventing path data recording sensors' occlusion, and these images have been employed in many successful path prediction methods. After collecting these images, their textual data are usually extracted. An efficient model must be able to perform prediction in any traffic condition and in a short time. However, the manual extraction of features for this purpose is difficult and time-consuming. Moreover, any error occurring in the first stage of feature annotation can lead to errors in the next stages and disrupt the entire performance. In this study, instead of manually defining features that express the traffic scene's content, we propose an image of the scene is given to the network at any time step t. We show that deep learning and CNNs can be an excellent alternative to manual feature extraction. We will delineate the proposed algorithm below.

## 3. Method

### 3.1. Problem Statement

To facilitate path programming and prevent accidents for realizing a fully automated driving, we require a simple yet powerful prediction framework to analyze the complex temporal dynamism of the participants in traffic in real-time.

Previous studies have assumed that the data received on each scene are first processed, and then, their local and temporal information is extracted. This information is used as the input to the prediction model. In this process, some information may be lost, and some extra information may be extracted.

An alternative method is to predict the future space for driving at a level close to the sensor's information, which needs lower levels for compressing the data [32]. Thus, it is better to directly select the data obtained from the sensor or the camera as the input to the model. We assume that a camera records the driving scene and sends the images to the prediction model to expect future positions based on these images for specified time intervals. This assumption has numerous merits: First, the training data are obtained by using a view of the driving scene easily and while avoiding the cumbersome process of data labeling; second, this approach resolves the need for features that are manually extracted.

Fig. 1 depicts the system model for the proposed structure. A set of observed images I and a set of future target coordinates C are assumed. It is assumed that all the frames can be obtained regularly, and h images are always accessible. The movement history of the vehicle is specified by a set of frames $I = \{i_{t1}, \dots i_{th}\}$ in the past th frames. Based on the sampling frequency of 25 Hz, the distance between every two sequential frames is 0.04 s.

We adopt the relative coordinate system, such that the TV in the current frame would be located on (0,0). If the TV's relative coordinates per frame are shown with $(X_t, Y_t)$, the model output is the estimation of the set, $C = \{(X_t, Y_t), \ldots, (X_{tf}, Y_{tf})\}$ in which tf is the number of time steps predicted by the model. The x- and y-axes demonstrate the vehicles' movement and the perpendicular direction, respectively.

The network receives a sequence of the vehicle's images and its SVs' path and generates a sequence of the vehicles' future positions.

A significant problem in designing prediction models is that the model should generalize from the training data. Since our model's input only consists of aerial images taken of the road and does not need any extra information, it can be used for any road.

### 3.1.1. Probabilistic Motion Prediction

Deep neural networks generate Gaussian distribution's parametric values to consider interacting vehicles' relevance [35]. This model estimates the probabilistic distribution of the vehicle's future positions, conditional to its movement history.

$$P(C|I) = p((X_t, Y_t), \ldots, (X_{tf}, Y_{tf}) | i_{t1}, \ldots i_{th})$$

The distribution of the time positions $(t, \ldots, t_f)$ can be determined by a bivariate Gaussian distribution with parameters $(\mu_t; \Sigma t)$.

$$C_t \sim N(\mu_t, \Sigma t)$$

where $\mu_t$ and $\Sigma t$ are the mean vector and covariance matrix of the Gaussian function of time t, respectively.

### 3.2. Network Architecture

Examining recent approaches shows that the path prediction algorithms' data have been manually labeled or obtained after extensive processing. The proposed method aims to reduce the manual labeling efforts of the input data and uses deep learning for real-time prediction on large driving datasets to develop the most current methods. Conv inspires our model. The social pooling model uses the SVs' information to generate the output. This pooling structure is similar to the method accomplished by Deo and Trivedi [11]; however, instead of using the features extracted from the SVs, we utilized the road image at any moment for training our model.

Due to the constraints and the necessity of temporal precision in a real-time system, and the need for easier debugging, we used a set of simple CNNs for encoding the information received from the camera. The encoder includes two convolutional layers and one pooling layer for each part of the image so that the size of the input image would be reduced, and its features would be extracted. There is one ReLU layer after each convolution layer. Each CNN processes one part of the I input sequence images with length h and summarizes it in a vector. Then, the extracted feature vectors are placed in a 3D tensor next to each other, thus creating a social tensor, so that the model can access the movement status of the vehicles and the spatial information of the scene. Finally, after the social tensor passes a CNN, an LSTM decoder receives the content vector containing the information obtained about the vehicles in the neighborhood and creates distribution parameters for the TV's predicted future positions (Fig. 2).

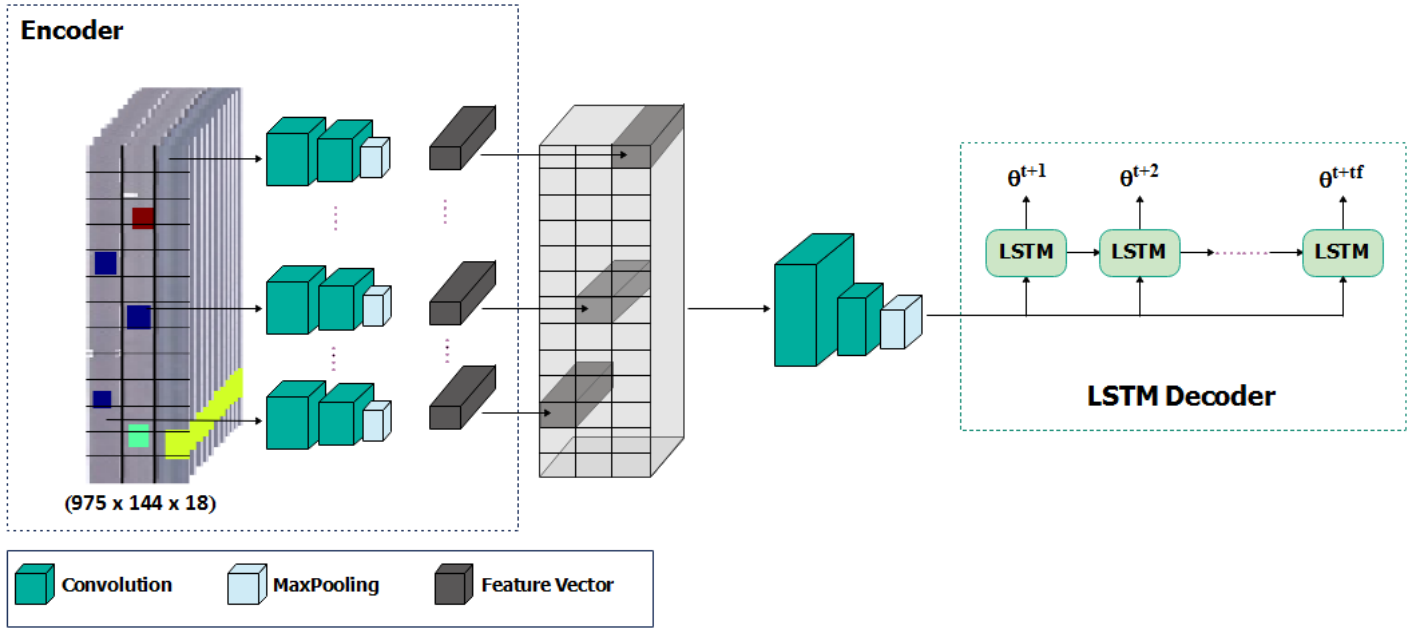

Fig. 2. The proposed model. The input images are fed to the network in a partitioned and sequential manner. Each part passes a $CNN_i$. The social pooling module learns the dependency between the vehicles. Finally, the LSTM decoder produces the distribution predicting the future path.

### 3.2.1. Training the feature extractor

Since the network input is path data and the path data are, in fact, a sequence, the network will have no accurate understanding of them by receiving images individually. Therefore, a sequence of images that expresses the vehicles' movement history is regarded as the input, the spatial features of which should first be extracted.

Today, the most popular feature detection tool is supervised learning, in which features start with random values and are gradually filtered using the machine learning algorithm. Such an approach requires appropriate training data and the precise selection of the parameters, mostly through trial and error.

CNNs have revolutionized pattern recognition [33]. Their main advantage is that the features are automatically learned from the training samples. The CNN approach is specifically powerful in image detection because the architecture of CNNs explicitly assumes that their inputs are images [34].

The proposed model receives an aerial view of the road with the TV in the center and a fixed region around it as the input. This region is regarded as the TV's neighborhood, and the effect of the vehicle's movement in this region on path prediction is examined. This neighborhood covers 180 feet in the longitudinal direction and the side lanes in the latitudinal direction.

In our method, the network should automatically extract the required information from input images, and spatial and temporal information was not explicitly given to the network. Therefore, to better understand the network, we divided the network into smaller parts, then each part will be processed separately.

Contrary to previous studies that have assumed only the SVs' features, we obtain the features of all the parts of the image, and the network finds the position of the vehicles and their movement pattern by receiving a sequence of input images. To access each part of the image, a grid area with the size MXN is assumed on the image. Here, N = 3 and M = 13 are assumed. For the jth part of the grid image, $S_j^{(i)}$, that is a sequence of the last h frames, $S_{j_t}^{(i)}, \ldots, S_{j_{th}}^{(i)}$, is sent to the CNN encoder as the input, such that $i \in I$ and $j \in (1, N * M)$.

As illustrated in figure 2, each part of the image passes a CNN so that its dimensions would be reduced, and its features would be extracted. It is noteworthy that the same parameters have been used for the convolution layer of all parts. Each CNN includes two convolutional layers. Convolution enables the model to learn the spatial dependencies. We used a 5 x 5 kernel with stride = 2 in each convolutional layer for recording the dependency of image components. There is a ReLU activation function after each of these layers. The final layer is a 2 x 2 max-pooling layer with stride = 2, which reduces the obtained feature map dimensions while maintaining the spatial features. After extracting all parts of image features, this obtained information was transferred to a social tensor to find the parts' relationship. The obtained features are in the form of a 2 x 4 x 8 tensor; for converting them into an input appropriate for the social tensor, they are flattened and converted into a 1D tensor with 64 elements.

The paths are separated into 8-sec parts, such that 3 sec would be regarded as the movement history, and the network should predict the next 5 sec. To reduce the complexity, and since sequential frames do not have additional information over each other, previous images were downsampled with a factor of 4 before being fed to the network.

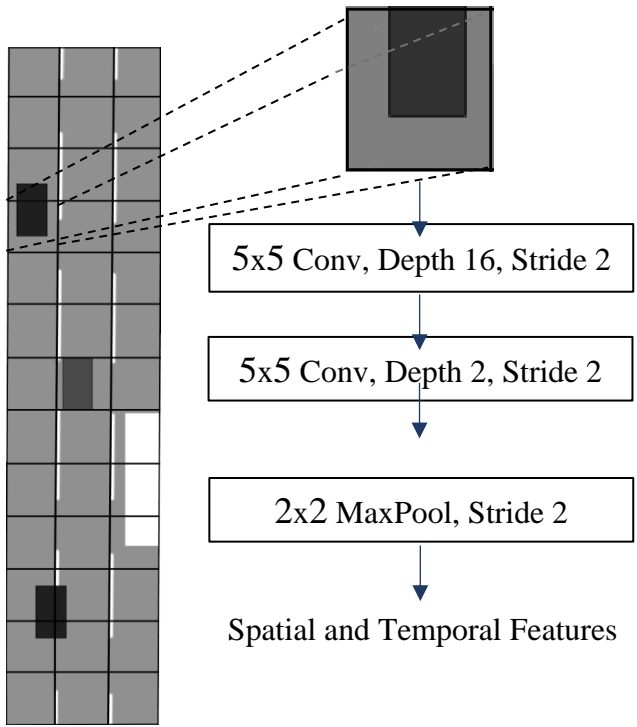

Figure 3. CNN encoder. The input image is segmented into a part of 13×3. Each part's features are extracted by passing a CNN consisting of two convolutional layers and one pooling layer.

### 3.2.2. Attention Module

Since the vehicles' behavior on the road is highly dependent on one another, it is essential to consider the interconnections between vehicles. Attention module is used for registering spatial and temporal dependency in long distances and models the relationships between the different parts of the road in the grid. Instead of modeling the relationships between vehicles per time step, the feature vectors extracted by the CNN encoder obtained from the time sequences are used as the inputs to the social tensor.

If the size of the encoded vectors of the path is S, the attention module with dimensions $N \times M \times S$ is considered.

For each part of the input image, a 64-unit vector is extracted by the convolutional layers that possess the vehicles' spatial and temporal features. These vectors are placed next to one another to form a 3D tensor with dimensions $3 \times 13 \times 64$. Their location is proportional to the primary partitioning of the image. In this section, the model achieves an appropriate understanding of the scene because it has access to the movement status and spatial order of the scene's components. Since the network should distinguish near and far neighbors from each other, convolutional layers and pooling after the social tensor are used. The convolutional layers' equation can help learn the useful local features inside the social tensor, and the max-pooling layer is utilized for adding local transitional invariance [11]; finally, the coded social content is obtained.

### 3.2.3. LSTM Decoder

Due to the complexity of traffic scenes and differences in driving styles, different vehicles' driving trajectories are mostly probabilistic in traffic. To show the probabilities in future trajectories, we used an LSTM-based decoder in order to generate the prediction distribution of future positions on target vehicles in the future frame of $t_f$. The decoder receives a content-based vector including an abstract of the information about different components of the road and dependency between them and generates the parameters of conditional distribution $P_\Theta(C|I)$ for future conditions of the target vehicle for time steps of $t = t+1, \ldots, t+t_f$. Where $\Theta$ indicates the Gaussian mixture model (GMM) parameters.

$$\Theta = (\Theta^{t+1}, \Theta^{t+2}, \ldots, \Theta^{t+tf})$$

Such that every LSTM can generate a $\Theta t$ related to the corresponding frame in the future.

### 3.2.4. Training and implementation details

We trained our model by using Adam's optimization function with a learning rate of 0.001. The batch size is equal to 1. In the decoder part, LSTMs with the dimensions of 128 were used. The ReLU activator function with α = 0.1 was used in all the layers. The model was executed in a GPU=NVIDIA GeForce GTX 1050Ti 4GB and RAM=8.0GB by using the PyTorch framework.

### 4. Dataset and Features

#### 4.1. HighD Dataset

The HighD data set [35] is recorded in 147 hours, including 44500 km of driving in six places near Cologne, Germany. The places differ in terms of the number of lanes and speed limitations. The recordings include light and heavy traffic and more than 110500 vehicles.

First, by using a drone, images of the road surface from the aerial view were taken, and the path of each vehicle (type, size, and maneuver) was extracted.

This dataset contains 60 recorded files, each containing 17 min of driving on average. This data set's scope covers part of the road with a length of 420 m, and each vehicle is visible for 13.6 sec on average. The videos are taken with a resolution of 4k, and their pixel precision is 25 frames per second [36].

**4.2. Image input representation**

In most studies on path prediction and problems related to traffic, textual databases are used. After imaging the road by executing the algorithms and performing computations, the images are annotated, and their information is extracted to collect this database. The direct use of the road images as the input to the prediction models can eliminate the time-consuming and challenging information extraction process.

To this end, we prepared simulated aerial images by using the information in the HighD database. The overall process of creating the images is depicted in Fig. 3.

| **Generate image using excel file information** |
|---|
| Inputs: Coordinates, Dimensions, ID, and Velocity of Vehicles [(x,y), (w,h), ID, V] |
| Output: Top View Image |
| Read Excel File <br> [(x,y), (w,h), ID, V] ← [(x,y), (w,h), ID, V] /= 0.10106 <br> [($x_t$,$y_t$), ($w_t$,$h_t$), $ID_t$, $V_t$] ← Target vehicle[(x,y), (w,h), ID, V] <br> If $V_t$ > 0 <br>    F1 ← Target First Frame <br>    F2 ← Target Last Frame <br>    For F1 < F < F2 <br>       back ← $x_t$ - ((90 * 0.3)/0.10106) <br>       front ← $x_t$ + $w_t$ + ((90 * 0.3)/0.10106) |

```
        left  ← (y_t - 6 )/0.10106
        right ← (y_t + h_t + 6)/0.10106
        ROI ← [back:Front , left:right]
        N ← number of vehicles in ROI
        Img ← Plot rect [(x_t , y_t) , (w_t , h_t)]
        For 1< i < N
           If (x_i , y_i) in ROI
              Img ← Plot rect [(x_i , y_i) , (w_i , h_i)]
           end
        end
        Image ← croped ROI from img
        Save Image
      end
  end
```

This procedure is iterated for all vehicles in the dataset. As mentioned before, the HighD dataset was saved in 60 separate files; in each, there were three Excel files containing information about the vehicles and one image of a road, on which the database was collected. To generate our dataset, we used the information in one of these files, including speed, ID, lane number, frame, and the longitudinal and latitudinal coordinates of the vehicles, and the corresponding aerial shot was selected as the background of our database. The origin of these images' coordinate system was located in the upper left-hand corner of the image, with the positive direction to the right and down for x- and y-axes, respectively. The Excel files' information can be accessed for two carriageways, but we selected only the vehicles that moved in the positive direction of the x-axis to create our dataset. It is noteworthy that information such as the vehicle's position or size was per meter in the database, and we mapped them to the pixel size.

In the next stage, we selected a vehicle as to the TV. Like human drivers that do not pay equal attention to all the vehicles on the road, we specified the neighborhood with the TV being the center to demonstrate the higher importance of closer vehicles than farther vehicles. The vehicles on the left and right lanes and within six vehicles in front and six vehicles behind the TV were regarded as the TV's neighborhood. We drew the vehicles in this region, such that instead of each vehicle, a rectangle proportionate to the size of the vehicle

(in random color) was placed in its $(x, y)$ position. The $(x, y)$ coordinates per frame indicated the upper left-hand corner of the vehicle.

Then, this region was cropped from the image and saved. This would be performed from the frames in which the TV entered the image to the frame where it existed the image. Fig. 4 displays an example of the created image.

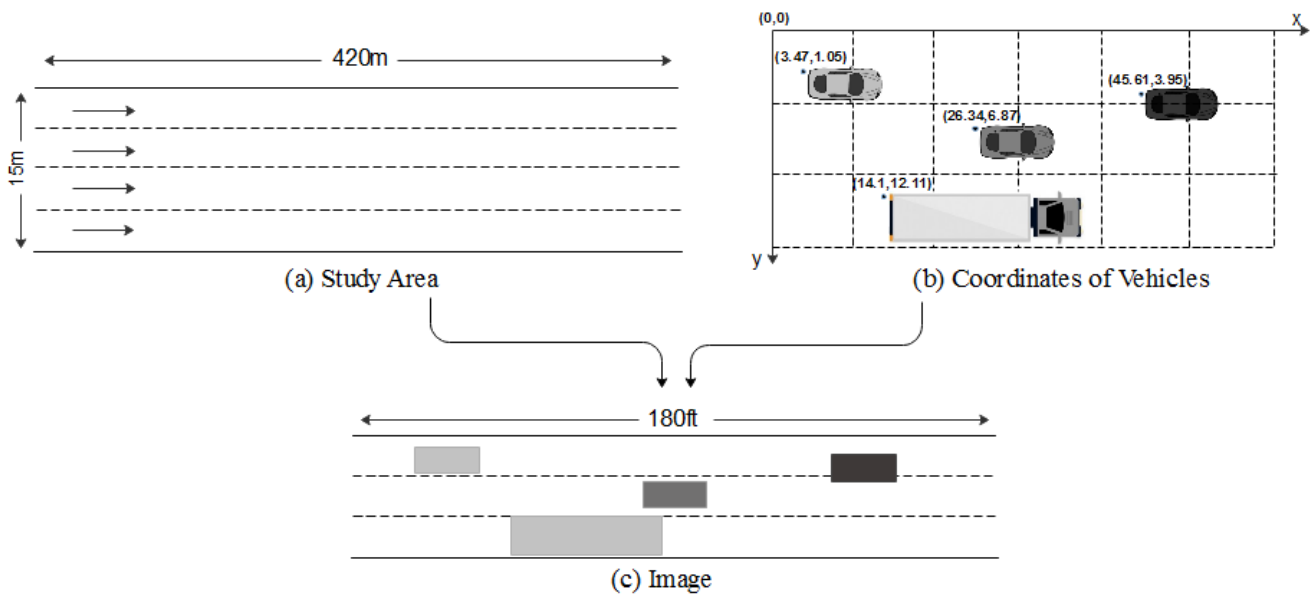

Figure 3. The method of creating an input image. a) main road image, b) the coordinates of the vehicles in the road, c) it was created a top view image with the target vehicle center and a vicinity of 180 feet length and corner lanes in both sides of the TV

The visible paths for the 420 TVs from the moment of entering the road to the moment of exiting the road, while taking into account the neighborhood, were saved in 420 folders. The images were classified into three categories: Training, test, and evaluation, with 60% percent of the images belonging to the training set, 20% to the test set, and 20% to the evaluation set.

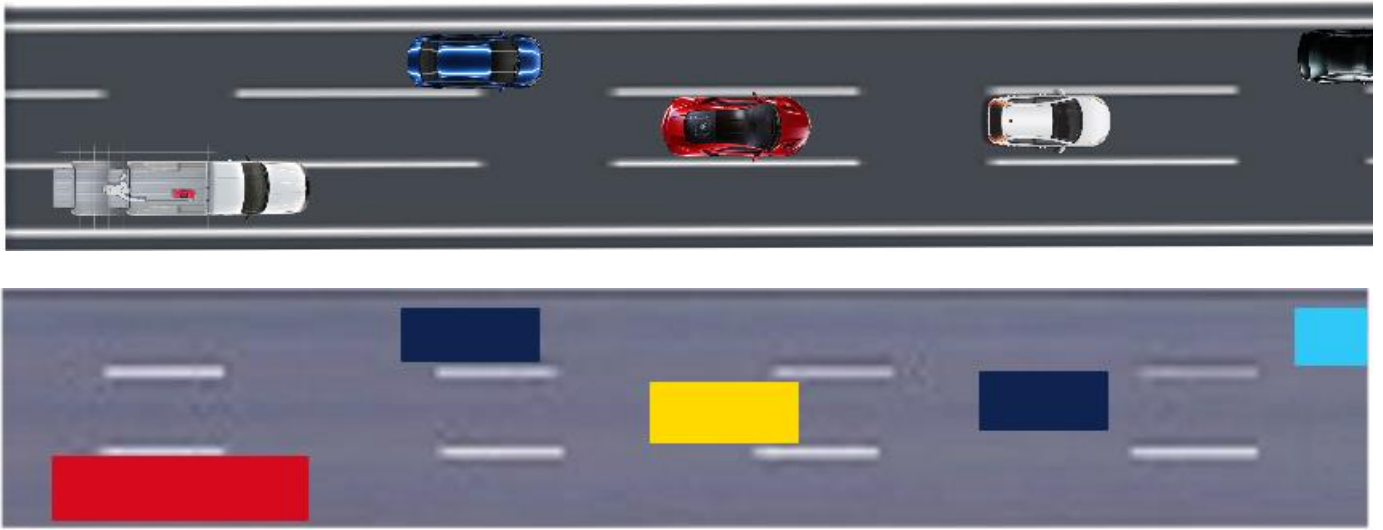

Fig. 4, Top: Original image of the road; Bottom: Simulated image; The yellow rectangle shows the TV.

## 5. Experimental Evaluation

### 5.1. Evaluation metric

In this project, the predicted paths' RMSE criterion based on the correct future paths was used for evaluating the proposed model. RMSE is a measurement criterion mostly used for measuring the difference between the values predicted by a model and the actual output values. It is an appropriate and tangible criterion for examining the precision of model predictions. RMSE is calculated as follows:

$$RMSE = \sqrt{\frac{(xpred - x)^2 + (ypred - y)^2}{N}}$$

Where (x,y) is the actual output coordinates, (xpred,ypred) is the coordinates predicted by the network, and N is the number of regions in the image.

For LSTM models that generate bivariate Gaussian distributions, the Gaussian distribution parameters such as μX, μY, σX, σY, and ρ are used for calculating RMSE. $xpred = μX$ and $ypred = μY$.

$$RMSE = \sqrt{\frac{(μX - x)^2 + (μY - y)^2}{N}}$$

### 5.2. Models Comparison

A comparison is performed on the following models, all of which consider the interaction between the SVs. They first receive the TV and SVs path as the input and produce the distribution predicting the future coordinates of the TV.

Social LSTM (S-LSTM) [10]: It is a social encoder-decoder model using a fully connected layer.

Convolutional Social Pooling (CS-LSTM) [11]: It is a social encoder-decoder using convolutional pooling.

CS-LSTM (M) Convolutional Social Pooling (maneuver-based) [11]: It is a complete model of CS-LSTM that creates multi-state road prediction based on six maneuvers (two longitudinal and three lateral).

[37] Non-local Social Pooling (NLS-LSTM): It combines local and non-local operations to produce an adaptable content vector for social pooling.

### 5.3. Quantitative Results

Table 1 presents the RMSE values for the compared models on the HighD database. Previous studies [3, 4] have compared their results with environment-independent prediction models to demonstrate the importance of considering surrounding factors. Here, we show that considering the SVs is a significant factor for path prediction and emphasizes the model's input. CS-LSTM and CS-LSTM(m) obtained better results than S-LSTM, showing the advantage of using convolution layers for modeling the dependency of vehicles' movement. Moreover, CS-LSTM(m) has a smaller error because it considers the future path's multi-state nature [11]. Our method obtained a smaller error on seconds 4 and 5 than the other methods. Concerning the low computational complexity of the proposed method, this can indicate its success in path prediction.

Table 1. The RMSE values in 5 sec predicted for the compared models

|      | Time horizon(s) | S_LSTM | CS_LSTM | CS_LSTM(M) | NLS-LSTM | OURs |
|------|-----------------|--------|---------|------------|----------|------|
| RMSE | 1               | 0.22   | 0.22    | 0.23       | **0.20** | **0.**42 |
|      | 2               | 0.62   | 0.61    | 0.615      | **0.57** | 0.88 |
|      | 3               | 1.27   | 1.24    | 1.29       | **1.14** | 1.26 |
|      | 4               | 2.15   | 2.10    | 2.18       | 1.9      | **1.57** |

| | 5 | 3.41 | 3.27 | 3.31 | 2.91 | **1.91** |

In figure 4, the value of RMSE in terms of the meter is reported for each of the time steps in the future for each model.

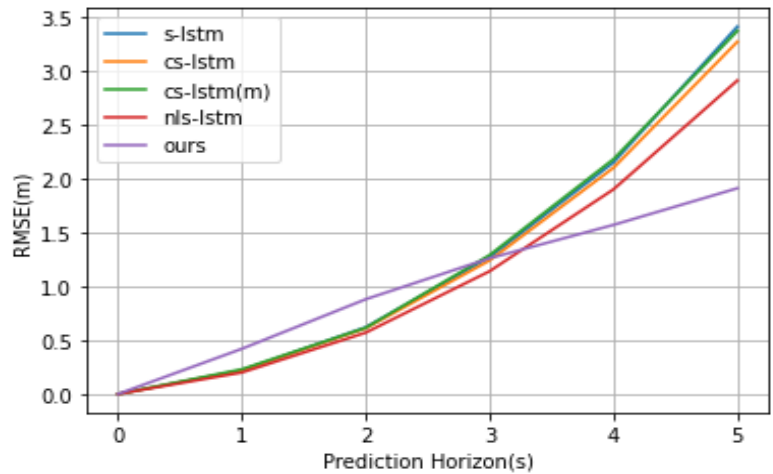

Figure 4. Comparison of quantitative results of different models

## 5.4. Qualitative Results

We display the results of the decoder output on our test set. Fig. 5 illustrates the history of the movement path of a vehicle and its predicted path. For the predicted path, a 31-unit set of tuples in the next 5 sec is produced. Each tuple demonstrates the Gaussian bivariate distribution parameters per frame. In the following figure, red dots show the predicted coordinates, while green dots indicate the actual output. The white lines show the track history in the past 3 sec. We can see that the uncertainty for longitudinal prediction is higher than the lateral position, which is reasonable concerning the road's geometry and the traffic flow.

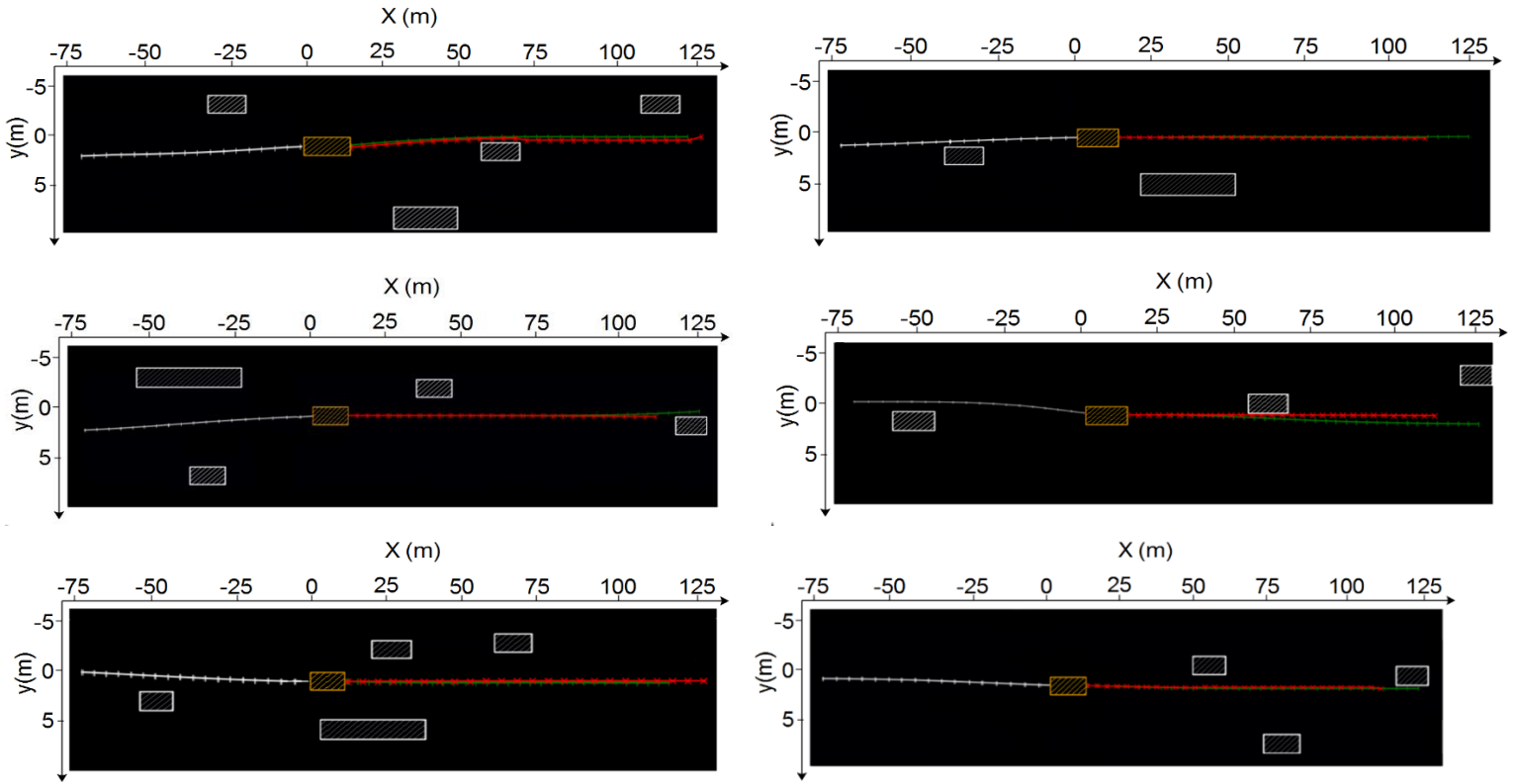

Figure 5. Predicted trajectories of TV. The yellow lines and white lines show the TV and TV movement history, respectively. The green and the red lines indicate ground truth and the predicted trajectories by model, respectively.

## 6. Conclusions and Future Work

The proposed algorithm's goal was to predict the path of the vehicles by using deep neural networks while presenting a model with minimum complexity. The necessity of this study stems from the importance of maintaining safety in driver-assistance systems and automated vehicles. The existing algorithms use large databases that contain a large volume of information, and most of them are networks with computational complexities. Therefore, we proposed a model that can make the prediction only by receiving images from the road's aerial image while having low computational complexity.

To solve this problem, we utilized deep learning methods that used convolutional layers for extracting the input images' features, employed social tensors for understanding the relationships between vehicles, and, finally, generated the future path sequence via an LSTM-based decoder.

We experimentally showed that the CNNs could learn the overall movement pattern of the vehicles. CNNs can learn the significant road features from a simple top-to-down image. One limitation of this model was in using the simulated images and the small size of the database. A future study can focus on utilizing a larger number of actual images for training the model.

It was also shown that the data belonging to the drones presented explicit information for extracting the path of vehicles and had to be examined, as suggested by previous studies.

# References

[1] A. Lawitzky, D. Althoff, C. F. Passenberg, G. Tanzmeister, D. Wollherr, and M. Buss, "Interactive scene prediction for automotive applications," in *2013 IEEE Intelligent Vehicles Symposium (IV)*, 2013, pp. 1028–1033.

[2] Z. Li, "Prediction of Vehicles' Trajectories based on Driver Behavior Models," 2014.

[3] S. Ulbrich and M. Maurer, "Towards tactical lane change behavior planning for automated vehicles," in *2015 IEEE 18th International Conference on Intelligent Transportation Systems*, 2015, pp. 989–995.

[4] S. Sivaraman and M. M. Trivedi, "Dynamic probabilistic drivability maps for lane change and merge driver assistance," *IEEE Trans. Intell. Transp. Syst.*, vol. 15, no. 5, pp. 2063–2073, 2014.

[5] N. Deo and M. M. Trivedi, "Multi-modal trajectory prediction of surrounding vehicles with maneuver based lstms," in *2018 IEEE Intelligent Vehicles Symposium (IV)*, 2018, pp. 1179–1184.

[6] S. Jawed, E. Boumaiza, J. Grabocka, and L. Schmidt-Thieme, "Data-driven vehicle trajectory forecasting," *arXiv Prepr. arXiv1902.05400*, 2019.

[7] Ü. Dogan, J. Edelbrunner, and I. Iossifidis, "Autonomous driving: A comparison of machine learning techniques by means of the prediction of lane change behavior," in *2011 IEEE International Conference on Robotics and Biomimetics*, 2011, pp. 1837–1843.

[8] T. Hirakawa, T. Yamashita, T. Tamaki, and H. Fujiyoshi, "Survey on vision-based path prediction," in *International Conference on Distributed, Ambient, and Pervasive Interactions*, 2018, pp. 48–64.

[9] G. Raipuria, "Vehicle Trajectory Prediction Using Road Structure," 2017.

[10] A. Alahi, K. Goel, V. Ramanathan, A. Robicquet, L. Fei-Fei, and S. Savarese, "Social lstm: Human trajectory prediction in crowded spaces," in *Proceedings of the IEEE conference on computer vision and pattern recognition*, 2016, pp. 961–971.

[11]   N. Deo and M. M. Trivedi, "Convolutional social pooling for vehicle trajectory prediction," in *Proceedings of the IEEE Conference on Computer Vision and Pattern Recognition Workshops*, 2018, pp. 1468–1476.

[12]   S. Lefèvre, D. Vasquez, and C. Laugier, "A survey on motion prediction and risk assessment for intelligent vehicles," *ROBOMECH J.*, vol. 1, no. 1, p. 1, 2014.

[13]   S. Mozaffari, O. Y. Al-Jarrah, M. Dianati, P. Jennings, and A. Mouzakitis, "Deep learning-based vehicle behaviour prediction for autonomous driving applications: A review," *arXiv Prepr. arXiv1912.11676*, 2019.

[14]   D. Yus, "Long-term vehicle movement prediction using Machine Learning methods." 2018.

[15]   A. Zyner, S. Worrall, and E. Nebot, "A recurrent neural network solution for predicting driver intention at unsignalized intersections," *IEEE Robot. Autom. Lett.*, vol. 3, no. 3, pp. 1759–1764, 2018.

[16]   A. Zyner, S. Worrall, J. Ward, and E. Nebot, "Long short term memory for driver intent prediction," in *2017 IEEE Intelligent Vehicles Symposium (IV)*, 2017, pp. 1484–1489.

[17]   L. Xin, P. Wang, C.-Y. Chan, J. Chen, S. E. Li, and B. Cheng, "Intention-aware long horizon trajectory prediction of surrounding vehicles using dual lstm networks," in *2018 21st International Conference on Intelligent Transportation Systems (ITSC)*, 2018, pp. 1441–1446.

[18]   D. J. Phillips, T. A. Wheeler, and M. J. Kochenderfer, "Generalizable intention prediction of human drivers at intersections," in *2017 IEEE Intelligent Vehicles Symposium (IV)*, 2017, pp. 1665–1670.

[19]   J. Li, B. Dai, X. Li, X. Xu, and D. Liu, "A dynamic bayesian network for vehicle maneuver prediction in highway driving scenarios: Framework and verification," *Electronics*, vol. 8, no. 1, p. 40, 2019.

[20]   S. Dai, L. Li, and Z. Li, "Modeling vehicle interactions via modified LSTM models for trajectory

prediction," *IEEE Access*, vol. 7, pp. 38287–38296, 2019.

[21]  F. Altché and A. de La Fortelle, "An LSTM network for highway trajectory prediction," in *2017 IEEE 20th International Conference on Intelligent Transportation Systems (ITSC)*, 2017, pp. 353–359.

[22]  M. S. Shirazi and B. T. Morris, "Looking at intersections: a survey of intersection monitoring, behavior and safety analysis of recent studies," *IEEE Trans. Intell. Transp. Syst.*, vol. 18, no. 1, pp. 4–24, 2016.

[23]  X. Li, X. Ying, and M. C. Chuah, "Grip: Graph-based interaction-aware trajectory prediction," in *2019 IEEE Intelligent Transportation Systems Conference (ITSC)*, 2019, pp. 3960–3966.

[24]  F. Diehl, T. Brunner, M. T. Le, and A. Knoll, "Graph neural networks for modelling traffic participant interaction," in *2019 IEEE Intelligent Vehicles Symposium (IV)*, 2019, pp. 695–701.

[25]  W. Ding and S. Shen, "Online vehicle trajectory prediction using policy anticipation network and optimization-based context reasoning," in *2019 International Conference on Robotics and Automation (ICRA)*, 2019, pp. 9610–9616.

[26]  D. Lee, Y. P. Kwon, S. McMains, and J. K. Hedrick, "Convolution neural network-based lane change intention prediction of surrounding vehicles for ACC," in *2017 IEEE 20th International Conference on Intelligent Transportation Systems (ITSC)*, 2017, pp. 1–6.

[27]  N. Djuric *et al.*, "Motion prediction of traffic actors for autonomous driving using deep convolutional networks," *arXiv Prepr. arXiv1808.05819*, vol. 2, 2018.

[28]  H. Cui *et al.*, "Multimodal trajectory predictions for autonomous driving using deep convolutional networks," in *2019 International Conference on Robotics and Automation (ICRA)*, 2019, pp. 2090–2096.

[29]  D. Nuss *et al.*, "A random finite set approach for dynamic occupancy grid maps with real-time

application," *Int. J. Rob. Res.*, vol. 37, no. 8, pp. 841–866, 2018.

[30] S. Hoermann, M. Bach, and K. Dietmayer, "Dynamic occupancy grid prediction for urban autonomous driving: A deep learning approach with fully automatic labeling," in *2018 IEEE International Conference on Robotics and Automation (ICRA)*, 2018, pp. 2056–2063.

[31] M. Schreiber, S. Hoermann, and K. Dietmayer, "Long-term occupancy grid prediction using recurrent neural networks," in *2019 International Conference on Robotics and Automation (ICRA)*, 2019, pp. 9299–9305.

[32] N. Mohajerin and M. Rohani, "Multi-step prediction of occupancy grid maps with recurrent neural networks," in *Proceedings of the IEEE Conference on Computer Vision and Pattern Recognition*, 2019, pp. 10600–10608.

[33] A. Krizhevsky, I. Sutskever, and G. E. Hinton, "Imagenet classification with deep convolutional neural networks," in *Advances in neural information processing systems*, 2012, pp. 1097–1105.

[34] M. Bojarski *et al.*, "End to end learning for self-driving cars," *arXiv Prepr. arXiv1604.07316*, 2016.

[35] R. Krajewski, J. Bock, L. Kloeker, and L. Eckstein, "The highd dataset: A drone dataset of naturalistic vehicle trajectories on german highways for validation of highly automated driving systems," in *2018 21st International Conference on Intelligent Transportation Systems (ITSC)*, 2018, pp. 2118–2125.

[36] "The highD Dataset." [Online]. Available: https://www.highd-dataset.com/format. [Accessed: 26-Jun-2020].

[37] K. Messaoud, I. Yahiaoui, A. Verroust-Blondet, and F. Nashashibi, "Non-local Social Pooling for Vehicle Trajectory Prediction," in *2019 IEEE Intelligent Vehicles Symposium (IV)*, 2019, pp. 975–980.

[38] Moradipari A, Tucker N, Alizadeh M. Mobility-Aware Electric Vehicle Fast Charging Load Models

with Geographical Price Variations. IEEE Transactions on Transportation Electrification. 2020 Sep 21.

[39] Moradipari A, Alizadeh M. Pricing differentiated services in an electric vehicle public charging station network. In2018 IEEE Conference on Decision and Control (CDC) 2018 Dec 17 (pp. 6488-6494).

[40] Moradipari A, Thrampoulidis C, Alizadeh M. Stage-wise Conservative Linear Bandits. Advances in Neural Information Processing Systems, vol. 33. 2020.

[42] Moradipari A, Alizadeh M, Thrampoulidis C. Linear thompson sampling under unknown linear constraints. InICASSP 2020-2020 IEEE International Conference on Acoustics, Speech and Signal Processing (ICASSP) 2020 May 4 (pp. 3392-3396).

[43] Moradipari A, Alizadeh M, Thrampoulidis C. Linear thompson sampling under unknown linear constraints. InICASSP 2020-2020 IEEE International Conference on Acoustics, Speech and Signal Processing (ICASSP) 2020 May 4 (pp. 3392-3396).

[44] Ebrahimzadeh E, Shams M, Jounghani AR, Fayaz F, Mirbagheri M, Hakimi N, Rajabion L, Soltanian-Zadeh H. Localizing confined epileptic foci in patients with an unclear focus or presumed multifocality using a component-based EEG-fMRI method. Cognitive Neurodynamics. 2021 Apr;15(2):207-22.

[45] Heravi H, Delgarmi M, Jounghani AR, Ebrahimi A, Shamsi M. Depth Dataset Using Microsoft Kinect-v2. bioRxiv. 2021 Jan 1.

[46] Delgarmi M, Heravi H, Jounghani AR, Shahrezie A, Ebrahimi A, Shamsi M. Automatic Landmark Detection of Human Back Surface from Depth Images via Deep Learning. bioRxiv. 2021 Jan 1.

[45] Heravi H, Aghaeifard R, Jounghani AR, Ebrahimi A, Delgarmi M. EXTRACTING FEATURES OF THE HUMAN FACE FROM RGB-D IMAGES TO PLAN FACIAL SURGERIES. Biomedical Engineering: Applications, Basis and Communications. 2020 Dec 4;32(06):2050042.